\documentclass{article}
\usepackage[T1]{fontenc}
\usepackage{iclr2027_conference,times}
\usepackage{amsmath,amsfonts,bm}

\def\eqref#1{equation~\ref{#1}}
\def\1{\bm{1}}

\DeclareMathAlphabet{\mathsfit}{\encodingdefault}{\sfdefault}{m}{sl}
\SetMathAlphabet{\mathsfit}{bold}{\encodingdefault}{\sfdefault}{bx}{n}

\usepackage{amsmath,amssymb,mathtools,booktabs,multirow,graphicx,microtype,xcolor,enumitem,url,float,hyperref}
\usepackage[font=small,labelfont=bf,skip=3pt,justification=centering]{subcaption}
\definecolor{AccentColor}{HTML}{2A6363}
\hypersetup{colorlinks=true,citecolor=AccentColor,linkcolor=AccentColor,urlcolor=AccentColor}

\title{{\color{AccentColor}
Decoupling Token Roles \\ in Autoregressive Pretraining
}}

 \iclrfinalcopy
\author{Suqin Yuan\textsuperscript{1} \quad Runqi Lin\textsuperscript{2} \quad Kevin Qinghong Lin\textsuperscript{2} \quad Junchi Yu\textsuperscript{2} \\ \textbf{Lei Feng}\textsuperscript{3} \quad \textbf{Chris Russell}\textsuperscript{2} \quad \textbf{Tongliang Liu}\textsuperscript{1*} \\
\textsuperscript{1} University of Sydney \quad
\textsuperscript{2} University of Oxford \quad
\textsuperscript{3} Southeast University
}

\begin{document}
\maketitle
\raggedbottom

\maketitle
\begingroup
\renewcommand{\thefootnote}{}
\footnotetext{\textsuperscript{*}Corresponding author.}
\endgroup

\begin{abstract}
Autoregressive pretraining increasingly draws on heterogeneous data, making it important to understand how a model learns from an individual token. The next-token prediction objective naturally identifies a token's contribution with its own loss. However, each token is not only a prediction target but also context for what follows. Using controlled corruption, we decouple these two roles and find a reversal: making a noisy token easier to predict reduces its damage as a target but increases it as context. The same decoupling helps explain text generated by language models: generation selects each token by its fit to the prefix, while its role as context is never tested against an independently determined continuation, because that continuation is generated to fit it. At known corrupted positions, acting through the context can reduce damage that removing the token's own loss does not. Understanding and controlling what a model learns from a token therefore requires decoupling its roles.
\end{abstract}

\section{Introduction}
\label{sec:introduction}

Autoregressive pretraining underlies current frontier language models~\citep{radford2019language,brown2020language,openai2023gpt4,grattafiori2024llama3,deepseekai2024deepseekv3,yang2025qwen3} and draws on ever larger and more varied text. Scaling laws call for more training tokens as models grow~\citep{kaplan2020scaling,hoffmann2022training}, while high-quality human-written text is limited~\citep{villalobos2024will,muennighoff2023scaling}, so training corpora increasingly combine filtered web crawls~\citep{penedo2024fineweb,soldaini2024dolma} with text generated by language models~\citep{gunasekar2023textbooks,maini2024rephrasing,allal2025smollm2}. Deciding what enters training has therefore become as important as how much data to use, and such decisions are now made at every level, from the mixture of sources~\citep{xie2023doremi,liu2025regmix} and the filtering of documents~\citep{wenzek2020ccnet,wettig2024qurating,li2024datacomplm} down to individual tokens~\citep{lin2024rho1,hans2024goldfish,rathi2026shaping,fang2024longppl}. At that finest level, deciding what enters training requires knowing what a model learns from an individual token in autoregressive pretraining.

The next-token objective suggests a simple account. The model is trained to predict each token $x_i$ from its prefix $x_{<i}$, and the loss $\ell_i$ records how well it does so. Every loss term is attached to the token being predicted, so a token's contribution to training is naturally identified with its loss, and token-level methods that mask or reweight a position's loss act on this view~\citep{lin2024rho1,fang2024longppl,helm2025tokenweighting}. The same token, however, enters training a second time. Once predicted, $x_i$ becomes context for every later prediction, and the model learns to use it. The consequences of this use appear in the losses of later tokens, never at position $i$. A token therefore takes part in training in two roles: in its \emph{target role} the model learns to predict it, and in its \emph{context role} the model learns to use it. In ordinary training the two roles always come together, because the same token is both predicted and used, and a token's loss describes only its target role.

To separate the two roles, we follow the use of corrupted labels to study generalization~\citep{zhang2017understanding} and use controlled corruption. We replace a fraction of the training tokens and route the same corruption, at the same positions, only into the prediction targets, only into the context, or into both. We then compare the trained models on clean held-out text. Since a token's loss is all that the objective records for it, we choose two kinds of replacement that differ sharply in their \emph{own loss}, the loss at the noisy token's own position. \emph{Random replacement} draws the noisy token from the vocabulary, so it is hard to predict from its prefix. \emph{Predictable replacement} lets a small language model choose a noisy token that fits the same prefix, so it is much easier to predict.

\begin{figure}[t]
\vskip -0.4in
\begin{minipage}{0.35\textwidth}\centering\includegraphics[width=\linewidth]{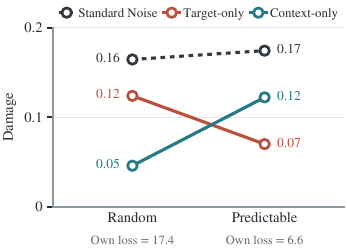}\end{minipage}\hfill
\begin{minipage}{0.6\textwidth}
\vskip +0.05in
\caption{Despite very different own losses, random and predictable replacement cause similar damage when the corruption enters both roles. When it enters one role at a time, predictable replacement causes less damage than random replacement as a target and more as context. Damage is the increase in held-out cross entropy over clean training for GPT-2-large~\citep{radford2019language} trained on $9$B tokens of OpenWebText~\citep{gokaslan2019openwebtext}. Own losses are means over $128$ matched positions under the clean model.}
\vskip -0.1in
\label{fig:reversal}
\end{minipage}
\vskip -0.18in
\end{figure}

We measure damage as the increase in held-out cross entropy over clean training. As shown in Figure~\ref{fig:reversal}, \emph{Random} and \emph{Predictable} replacement cause similar damage when the corruption enters both roles, even though their own losses differ sharply. Controlled corruption allows us to route the same noisy tokens into one role at a time and compare the two replacements within each role. When the corruption enters only the targets, the damage follows own loss: predictable replacement causes less damage than random replacement. When it enters only the context, the order reverses: predictable replacement causes more. This reversal holds for SmolLM2~\citep{allal2025smollm2} trained from scratch and for continued pretraining of Qwen3~\citep{yang2025qwen3}, across model sizes, training budgets, corruption rates and seeds.

To understand the reversal, we consider the two relations every token has with its text: to the prefix from which it is predicted, and to the continuation that is predicted from it. With controlled corruption we vary the two separately, choosing replacements whose fit to the prefix and fit to the continuation vary independently (Section~\ref{sec:relations}). A better fit to the prefix lowers the damage as a target whether or not the continuation fits. As context, the same change lowers the damage when the continuation fits and raises it when it does not. The two roles thus respond to different properties of the same token. Understanding what a model learns from a token therefore requires decoupling its roles: the same token can serve training well as a target and poorly as context, or the reverse.

This understanding extends to text generated by a language model, which is also a form of predictable replacement: every generated token is chosen by its fit to the prefix, and \emph{no token ever has to fit its future, because its future is generated to fit it}. We let a frozen model continue human text for $32$ tokens and then return to the original human continuation, fixing what follows independently of the generated text. The more greedily the model decodes, the easier the generated text is to predict and the harder the first human token after it is to predict. Even random tokens disturb the fixed continuation less than generated text does under any decoding. Using generated text for pretraining therefore turns predictability from an observed property of the data into one that the generation process itself selects for, even though increasing it can work against the token's context role in learning.

Decoupling the roles also shows where token-level control of learning should act. Token-level methods mask or reweight a token's loss, which acts only on its target role. Even when the corrupted positions are known exactly, removing their losses does not reduce the damage of predictable replacement, which is more damaging as context. A reliability flag that marks the same positions in the input acts on the context role instead and recovers about $40\%$ of the damage (Section~\ref{sec:interfaces}). Which tokens to act on and which role to act through are therefore separate decisions.

Our main contributions can be summarized as follows:
\begin{enumerate}[leftmargin=1.6em,itemsep=0.15em,topsep=0.25em]
\item Using controlled corruption routed into the targets, the context, or both, we show that making a noisy token easier to predict reduces its damage as a target and increases its damage as context.
\item Varying a noisy token's fit to its prefix and to its continuation separately, we show that damage as a target follows the fit to the prefix and that the continuation decides whether a better fit to the prefix lowers or raises the damage as context.
\item We show that autoregressive generation selects for the property that makes text easy to predict, while generating the continuation to fit those choices; when the continuation is fixed independently, increasing predictability makes the generated text fit it worse.
\item We show that token-level control should choose its role: knowing which tokens to act on does not determine whether to intervene through the target or context.
\end{enumerate}

\clearpage

\section{Related Work}
\label{sec:related-work}

\textbf{Two roles in the objective.}
Influence functions separate a training token's influence as an output from its influence as an input~\citep{grosse2023influence}, the gradient at a position has been split into a term from its own loss and terms routed back from later positions~\citep{liu2026alphatoken}, and state-prediction separation gives the two uses distinct representations~\citep{monea2026state}. Corrupting only the targets of a language model raises its loss far less than the corruption rate~\citep{peng2026robustness}, as does random noise in both roles~\citep{ru2025randomnoise}. Corrupting only the context, with clean targets, is a long-standing regularizer~\citep{xie2017datanoising,bowman2016generating,liu2022fcm,chen2026demystifying}, and scheduled sampling likewise alters the context while keeping the targets~\citep{bengio2015scheduled}, its uniform-replacement control comparing random with model-sampled context by generation quality; the objective it induces is improper~\citep{huszar2015hownot}. Replaced-token detection learns to find corrupted inputs~\citep{clark2020electra}, and translation studies separate source-side from target-side noise~\citep{zhu2024translationnoise}. Other work assigns tokens to one role by design: prefix language models and prompt-loss weighting keep some tokens as context only~\citep{wang2022architecture,shi2024instruction,huertaenochian2024promptloss}, and continual pretraining on noised text drops noised words as targets~\citep{kojima2025noisycontext}. Each of these acts on one role; comparing the two roles under the same corruption is what reveals the reversal.

\textbf{Scores at the token's own position.}
Token-level selection and weighting score the predicted token by its excess loss over a reference model~\citep{lin2024rho1,mindermann2022prioritized} or its predictive entropy~\citep{su2024mile}, truncate high-loss or high-error-norm tokens~\citep{kang2020loss,li2024ent}, or clean tokens in instruction data~\citep{pang2025tokencleaning}; document-level pruning aggregates the same quantities into a perplexity~\citep{wenzek2020ccnet,marion2023less,ankner2025perplexed}. Long-context scores compare the loss under a long and a short prefix~\citep{fang2024longppl,helm2025tokenweighting}, and context-length and attribution diagnostics measure how a prediction depends on the context it receives~\citep{khandelwal2018sharp,mohebbi2023valuezeroing,cohenwang2024contextcite}. All of these score a token by how it is predicted, that is, through its target role.

\textbf{Generated text.}
Truncated and greedy decoding concentrate on high-probability tokens~\citep{fan2018hierarchical,holtzman2020curious}, detectors exploit the resulting likelihood profile~\citep{gehrmann2019gltr,mitchell2023detectgpt}, and a model learns text written by another model more readily because of its lower perplexity~\citep{ren2024learnbetter}. Exposure-bias work feeds a model its own generated prefix and scores the continuation it then produces, which adapts to the prefix~\citep{ranzato2016sequence,he2021exposure}; later generated tokens are written to be consistent with earlier ones~\citep{zhang2024snowball}. We fix the human continuation. Synthetic pretraining data are screened by likelihood or by judges~\citep{gunasekar2023textbooks,maini2024rephrasing}; recursive training on generated data collapses the distribution unless real data are retained~\citep{shumailov2024collapse,gerstgrasser2024collapse}, more severely under narrow decoding~\citep{dohmatob2024tails,drayson2025detection}, and noise consistent with its sequence harms more than local noise~\citep{havrilla2024noise}. We separate what training learns from generated text as a target and as context.

\textbf{Loss masks and markers.}
Loss masks remove selected target losses and keep the text as context, to limit memorization~\citep{hans2024goldfish}, to understand high-risk text without generating it~\citep{wang2025slung}, or to remove capabilities~\citep{rathi2026shaping}; the masked identity remains encoded in the hidden states~\citep{kosireddy2026lossmasking}, and masking erroneous steps in synthetic reasoning data proves unnecessary when those steps are marked in the input~\citep{ye2024physics22}. Markers condition the model on a tag attached to a sentence, segment or document, identifying back-translated, undesirable, junk or hazardous text~\citep{caswell2019tagged,korbak2023pretraining,allenzhu2024physics33,lee2026mark}. We compare the two at the same oracle positions, where the loss mask acts on the target role and a reliability flag on the context role.

\section{The Two Roles Respond Differently to the Same Corruption}
\label{sec:roles}

In this section we introduce noisy tokens into the training text and route the same corruption separately through the prediction targets and the context. In ordinary training the two roles cannot be observed apart, because the same token is always both predicted and used. This lets us measure how the same corrupted tokens affect learning when the model is trained to predict them and when it is trained to use them. We find a reversal between the two roles: corruption that is easier to predict does less damage as a target but more as context. We then test how consistently it appears across training settings and which relation of the token, to its prefix or to its continuation, each role follows.

\subsection{Decoupling the two token roles}
\label{sec:protocol}

For a text $x=(x_1,\ldots,x_T)$, autoregressive pretraining minimizes~\citep{bengio2003neural,radford2018improving}
\begin{equation}
\mathcal{L}(\theta)=\sum_{i=1}^{T}\ell_i(\theta),
\qquad
\ell_i(\theta)=-\log p_\theta(x_i\mid x_{<i}).
\label{eq:ntp}
\end{equation}
For a token $x_k$, $\ell_k$ measures how well the model predicts it as a \emph{target} from its prefix. The same token also enters every later prediction through the \emph{context} $x_{<i}$, $i>k$. The objective thus contains both roles, but the token's own loss describes only the one in which it is predicted.

We therefore compare the training consequences of the two roles: what happens when the model learns to predict a token and what happens when it learns to use the same token as context. To make the two roles explicit, we write the objective with a \emph{context} stream $c$ and a \emph{target} stream $y$:
\begin{equation}
\mathcal{L}(\theta;c,y)=\sum_{i=1}^{T}-\log p_\theta(y_i\mid c_{<i}).
\label{eq:objective}
\end{equation}
Ordinary training sets $c=y=x$, which returns Equation~\ref{eq:ntp}. The notation only separates the two ways in which a token enters training; it does not tell us what consequence each has for learning. To measure this, we change the token in one role while leaving it unchanged in the other, and compare the resulting models. As \citet{zhang2017understanding} corrupted labels and inputs to study generalization, we corrupt the text in a known way, which lets us study these consequences under controlled conditions: we choose which positions change and what replaces them. In a classifier the labels and the inputs are distinct; in autoregressive training every token is both a label and an input. The simplest controlled change is to replace the token. Unlike deletion or insertion, replacement keeps the length of the text and the position of every other token, so a corrupted stream stays aligned with a clean one. Let $\tilde{x}$ be a corrupted copy of $x$ in which the tokens at a fixed set of positions are replaced. Since the two streams are written separately, the same corruption thus can enter either one while the other remains clean:
\begin{equation}
(c,y)=
\begin{cases}
(x,x), & \emph{Clean},\\
(x,\tilde{x}), & \emph{Target-only Noise},\\
(\tilde{x},x), & \emph{Context-only Noise},\\
(\tilde{x},\tilde{x}), & \emph{Standard Noise}.
\end{cases}
\label{eq:routing}
\end{equation}
In \emph{Target-only} training, a noisy token $\tilde{x}_k$ replaces the target at position $k$ but never enters the context, where the original $x_k$ remains. Only the prediction at position $k$ changes: the model learns to predict the noisy token from clean text. In \emph{Context-only} training, $\tilde{x}_k$ is never a prediction target, and $x_k$ is still predicted at position $k$. The noisy token becomes context for the predictions that follow: the model learns to predict clean text after it. \emph{Standard Noise} places the same noisy token in both roles, as ordinary training on corrupted text would. The three noisy conditions use the same corrupted positions and replacement tokens, together with the same training setup, so they differ only in the role through which the corruption enters.

\subsection{Comparing the two token roles}
\label{sec:comparison}

With the two roles decoupled, differences between noisy tokens can be examined separately in each role. We begin with prediction difficulty, the quantity that a token's own loss records and that token-level selection acts on, so that the response of each role can be compared with the score ordinary training already provides.
To compare tokens of very different difficulty at the same positions, we construct two corruptions that differ in how the replacement is chosen (Appendix~\ref{app:corruption}):
\begin{enumerate}[leftmargin=1.6em,itemsep=0.0em,topsep=0.0em]
\item \emph{Random replacement} draws uniformly from the vocabulary.

\item \emph{Predictable replacement} selects a highly ranked whole-word candidate from a frozen language model given the same clean prefix. 
\end{enumerate}

\begin{figure}[t]
\centering
\begin{subfigure}[t]{0.5\textwidth}\centering\includegraphics[width=\linewidth]{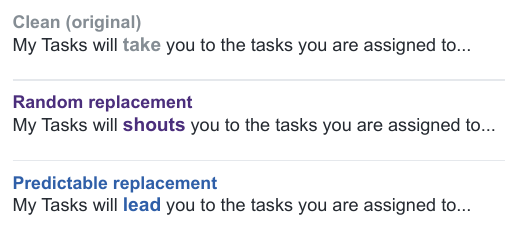}
    \vskip -0.02in
    \caption{Replacement examples}
\label{fig:32a}
\end{subfigure}\hfill
\begin{subfigure}[t]{0.45\textwidth}\centering\includegraphics[width=\linewidth]{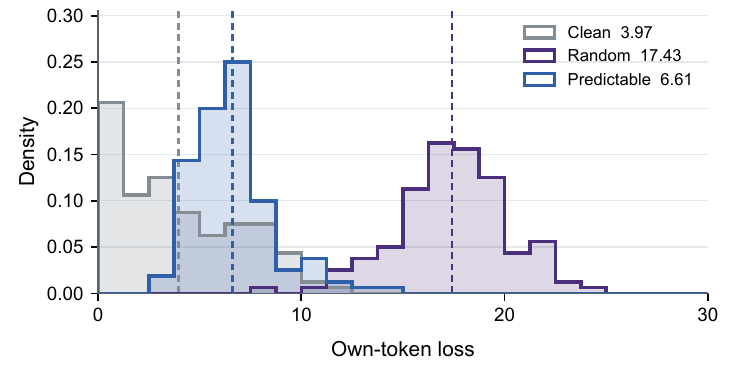}
    \vskip -0.02in
\caption{Own loss}
\label{fig:32b}
\end{subfigure}
    \vskip -0.05in
\caption{(a) One realized training position, with the same prefix and continuation for the original, random, and predictable tokens. (b) Own-loss distributions at $128$ matched positions under a fixed clean GPT-2-large model; dashed lines mark the means.}
\label{fig:32}
    \vskip -0.15in
\end{figure}

In both cases the surrounding text is left unchanged, and neither choice looks at the continuation. Figure~\ref{fig:32a} shows the two replacements at one realized position. Across $128$ matched positions, the mean own loss under a clean GPT-2-large model is $6.6$ for predictable replacement, $17.4$ for random replacement and $4.0$ for the original tokens (Figure~\ref{fig:32b}). Predictable replacements are therefore much easier to predict as targets from the same clean prefix, though still harder than the tokens they replace.

We then train each corruption under the four conditions in Equation~\ref{eq:routing}. We run this comparison with GPT-2-large~\citep{radford2019language} trained from scratch on OpenWebText~\citep{gokaslan2019openwebtext} using nanoGPT~\citep{Karpathy2022}. Both corruptions target a $10\%$ replacement rate, and within each corruption the same realized positions and replacement tokens are used across the three noisy routing conditions. The four conditions are trained under the matched setup above along the same $9$B-token schedule and evaluated on the same clean held-out set, and training details in Appendix~\ref{app:gpt2-setup}. 

\begin{figure}[h]
    \vskip -0.06in
\begin{minipage}{0.59\textwidth}
    \centering
    \includegraphics[width=8.1cm]{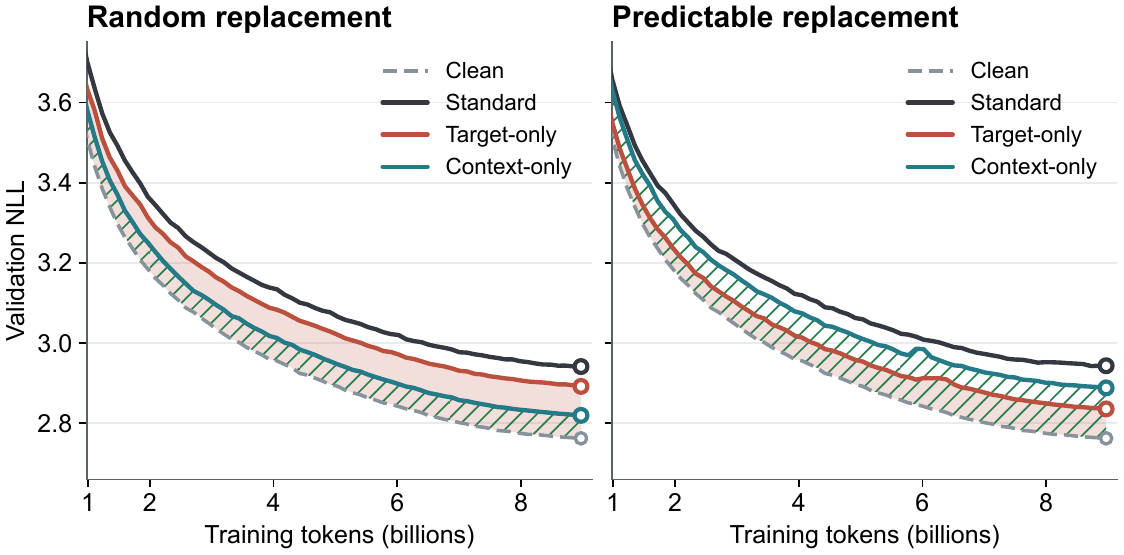} 
\end{minipage}
\begin{minipage}{0.40\textwidth} 
    \vskip -0.0in
    \centering
\caption{Throughout training, random and predictable replacement differ in which role is more damaging. Curves show held-out cross entropy during GPT-2 training, with damage given by the distance from the clean curve: target-only training is more damaging under random replacement, while context-only training is more damaging under predictable replacement.}
    \vskip -0.1in
    \label{fig1}
\end{minipage}
    \vskip -0.1in
\end{figure}

Figure~\ref{fig1} shows the held-out cross entropy of each condition over training, one panel per kind of replacement. We measure the consequence of each condition by the increase in clean held-out cross entropy relative to matched clean training, which we call \emph{damage}.
Despite the large difference in the noisy tokens' own losses, random and predictable replacement cause similar damage under Standard Noise (Figure~\ref{fig1}). Under target-only training, the difference follows own loss: predictable replacement causes less damage than random replacement. Under context-only training, however, the relation reverses: predictable replacement causes more damage.
Damage under context-only training comes from using the noisy tokens to predict the unchanged continuation. The next-token objective assigns the losses for these predictions to the later tokens, while the noisy tokens' own losses describe how well the noisy tokens themselves are predicted. Under predictable replacement, this context-side damage even exceeds target-side damage throughout training. The advantage that a lower own loss gives a noisy token as a target therefore does not extend to its role as context.

\begin{figure}[t]
\centering
    %\vskip -0.25in
\includegraphics[width=1.0\textwidth]{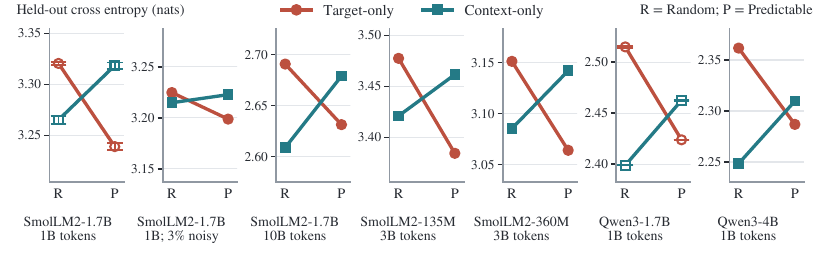}
    \vskip -0.15in
\caption{The reversal between the two roles is not specific to the initial GPT-2 comparison, and remains consistent across changes in model and training setup. Held-out cross entropy under target-only and context-only training for Random (R) and Predictable (P) replacement. The SmolLM2-1.7B and Qwen3-1.7B $1$B-token settings each show the mean and observed range over three seeds. 
}
\label{fig:settings}
    \vskip -0.15in
\end{figure}

\textbf{The reversal holds across settings.}
We repeat the comparison in two other systems. SmolLM2~\citep{allal2025smollm2} is trained from scratch on FineWeb-Edu~\citep{lozhkov2024fineweb-edu} with Nanotron~\citep{huggingface2023nanotron}, which changes the architecture, the training text and the training framework. Qwen3 base models~\citep{yang2025qwen3} are continued-pretrained on the same text with Hugging Face Transformers~\citep{wolf2020transformers}, where the corruption enters a model that is already pretrained. For SmolLM2 we also vary model size from $135$M to $1.7$B parameters, the training budget from $1$B to $10$B tokens, and the corruption rate from $10\%$ to $3\%$ (Figure~\ref{fig:settings}). Across these changes, predictable replacement remains less damaging than random replacement as a target and more damaging as context. At a $10\%$ rate, the difference is $0.06$ to $0.09$ as a target and $0.04$ to $0.07$ as context. Training for $10$B instead of $1$B tokens leaves both differences in place, and lowering the rate to $3\%$ shrinks them to $0.03$ and $0.01$ without changing their signs. We also repeat the SmolLM2-1.7B and Qwen3-1.7B $1$B-token settings with three seeds; the reversal remains in every run.

\subsection{Tracing the two token roles to the prefix and the continuation}

\begin{figure}[h]
    \vskip -0.15in
\centering
\begin{subfigure}[t]{0.435\textwidth}\centering\includegraphics[width=\linewidth]{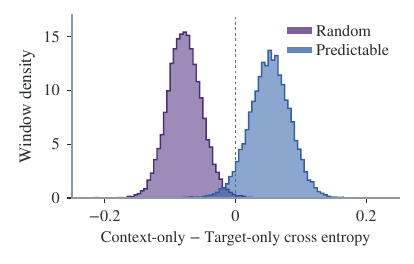}\caption{GPT-2-large, per held-out window}\end{subfigure}\hfill
\begin{subfigure}[t]{0.54\textwidth}\centering\includegraphics[width=\linewidth]{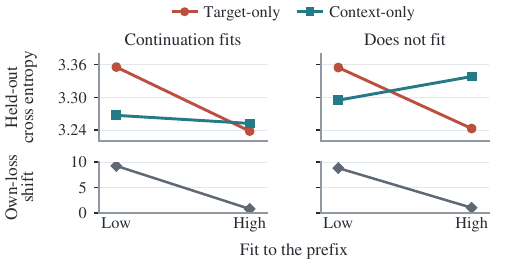}\caption{SmolLM2-1.7B, four replacement sets}\end{subfigure}
    \vskip -0.05in
\caption{(a) On nearly every held-out window, the context-only model is worse than the target-only model under predictable replacement and better under random replacement. Each value is the context-only minus the target-only held-out cross entropy of GPT-2-large on one of $19{,}531$ windows after $9$B tokens; positive values mean that the context-only model is worse. (b) A better fit to the prefix lowers target-only cross entropy whether or not the continuation fits, and lowers context-only cross entropy only when it fits (top row); the own-loss shift $\delta$ of Equation~\ref{eq:shift} falls in both cases (bottom row). The four SmolLM2-1.7B replacement sets pair a low and a high fit to the prefix with a continuation that fits (left) or does not (right).}
\label{fig:relations}
    \vskip -0.1in
\end{figure}

The comparison between the target-only and context-only models in Section~\ref{sec:comparison} holds on almost every held-out window. Across $19{,}531$ unseen GPT-2 windows, the context-only model is worse than the target-only model in $95.1\%$ of windows under predictable replacement and better in $99.6\%$ under random replacement (Figure~\ref{fig:relations}a).
Section~\ref{sec:comparison} located the context-side damage of predictable replacement in the predictions of the unchanged continuation. A replacement is predicted from its prefix and is used to predict its continuation. Random replacement controls neither relation, and predictable replacement is chosen for its fit to the prefix alone (Appendices~\ref{app:corruption} and~\ref{app:family-axes}), so the comparison leaves the fit to the continuation to chance. We therefore vary the two fits separately at the same positions.

\label{sec:relations}

\textbf{Varying the two relations.}
At each position, candidate replacements come from a common token-frequency group and are scored by a frozen language model on both relations: fit to the prefix by their probability under the clean prefix, and fit to the continuation by the loss on the clean continuation once the candidate is inserted. Four replacement sets pair a low and a high fit to the prefix with a continuation that fits and one that does not. We train each set under target-only and context-only training in the SmolLM2-1.7B setting of Figure~\ref{fig:settings} (Appendix~\ref{app:relation-map}).

A better fit to the prefix lowers the target-only cross entropy by about the same amount whether or not the continuation fits (Figure~\ref{fig:relations}b). The context-only model behaves differently: the same change lowers its cross entropy slightly when the continuation fits and raises it when the continuation does not. Making a replacement easier to predict thus has a stable effect in its target role, while its effect as context changes direction with the continuation.

Among the four sets, the reversal of Section~\ref{sec:comparison} thus appears only where the continuation does not fit. The comparison between the two models is the same in both columns: the target-only model is worse at low fit to the prefix and the context-only model at high fit, mainly because the damage as a target falls sharply as the replacement becomes easier to predict.

On the same sets we also measure the change in own loss when the clean target is replaced,
\begin{equation}
\delta_i=-\log p_\theta(\tilde{x}_i\mid x_{<i})+\log p_\theta(x_i\mid x_{<i}),
\label{eq:shift}
\end{equation}
under a fixed clean SmolLM2 checkpoint (Appendix~\ref{app:own-loss}). The shift falls with fit to the prefix by nearly the same amount in both columns (Figure~\ref{fig:relations}b, lower row). Own loss follows the fit to the prefix, as the target role does.

%\medskip
{\setlength{\fboxsep}{4pt}\noindent\fbox{\parbox{\dimexpr\linewidth-2\fboxsep-2\fboxrule\relax}{\textbf{Decoupling token roles.} The property that makes a token a better target can make it a worse context. Because every token plays both roles, their combined effect can hide what happens in each role, so understanding what a model learns from a token requires decoupling its roles.}}}
%\medskip

\section{Decoupling Token Roles in Practice}
\label{sec:consequences}
\label{sec:masking}
\label{sec:remedy}

In this section we apply the decoupling to two practical settings. We first consider text generated by a language model, where every token is chosen for its fit to the prefix, the same relation that makes it easy to predict as a target. We test whether this also makes the generated text reliable as context for what follows (Section~\ref{sec:generated}). We also use the two roles to examine how token-level interventions act when unreliable positions are known. We compare removing their losses with marking the same positions in the input, acting on the target and context roles respectively (Section~\ref{sec:interfaces}).

\subsection{From single corrupted tokens to generated text}
\label{sec:generated}

\emph{Predictable replacement} uses a language model to choose a token that is easy to predict from its prefix. Autoregressive generation does the same repeatedly: at every step, next-token probabilities select what is written next, producing text that the model itself finds exceptionally easy to predict. Generated text is therefore a kind of \emph{predictable replacement}. But this predictability comes entirely from the prefix, without regard to what comes after the generated text. Section~\ref{sec:relations} showed that making a token easier to predict does not have the same consequence in its two roles: it consistently helps as a target, while its effect as context depends on what follows. Generated text increasingly enters pretraining corpora~\citep{gunasekar2023textbooks,maini2024rephrasing,allal2025smollm2}, where these highly predictable tokens are both predicted and used as context.

We run the experiments with GPT-2 on OpenWebText and SmolLM2 on FineWeb-Edu, in a setup similar to Section~\ref{sec:comparison}. We select $10\%$ of non-overlapping stretches of $32$ consecutive tokens. For a stretch starting at position $k$, a frozen generator $p_\phi$ continues the human text,
\begin{equation}
\tilde{x}_t=\mathcal{D}\big(p_\phi(\cdot\mid \tilde{x}_{<t})\big),\qquad t=k,\dots,k+31,
\label{eq:generation}
\end{equation}
with $\tilde{x}_t=x_t$ at every other position, so that the original continuation resumes at $k+32$. The decoding rule $\mathcal{D}$ either takes the most probable token (\emph{greedy} decoding) or samples from the smallest set of tokens whose probability reaches $p$ (\emph{top-p} sampling). Each generated token is thus conditioned on the human prefix and the tokens generated before it; the continuation $x_{\ge k+32}$ never enters Equation~\ref{eq:generation}. A \emph{random} condition places random tokens at the same positions.

\begin{figure}[t]
\centering
\begin{subfigure}[t]{0.48\textwidth}\centering\includegraphics[width=\linewidth]{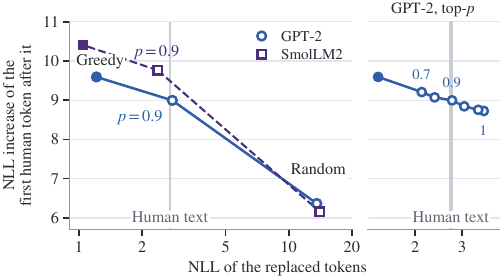}\caption{Fit to the prefix and to the continuation}\end{subfigure}\hfill
\begin{subfigure}[t]{0.43\textwidth}\centering\includegraphics[width=\linewidth]{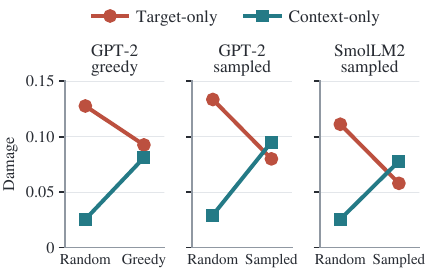}\caption{Damage by role}\end{subfigure}
    \vskip -0.1in
\caption{(a) Fluent generated text disturbs the first human token after it more than random tokens do. Each point places one kind of replacement by the NLL of the replaced tokens under the generator (fit to the prefix) and the increase in NLL of the first human token after them (fit to the continuation), averaged over $20{,}000$ fixed positions. Circles show GPT-2-medium on OpenWebText and squares SmolLM2-135M on FineWeb-Edu, each scored by its own generator; filled points are greedy decoding, and the band spans the mean NLL of the original human text under the two scorers. The right panel shows every top-p setting for GPT-2. (b) Relative to random tokens, generated text causes less damage under target-only training and more under context-only training. Each generated condition is paired with the Random and Clean runs of its own campaign.}
\label{fig:generated-spans}
    \vskip -0.15in
\end{figure}

Figure~\ref{fig:generated-spans}a places each replacement on the two relations of Section~\ref{sec:relations}: its fit to the prefix, measured by the mean NLL of the replaced tokens under the generator, and its fit to the continuation, measured by the increase in NLL of the first human token after them. The second relation is normally absent during generation. Each generated token is chosen from its prefix, but the tokens that follow are then generated from a prefix containing that choice. Generation therefore chooses both the token and the trajectory along which it will subsequently be used as context.
Our construction breaks this coupling after $32$ tokens: generation stops and the original human continuation resumes. The generated text must now serve as context for a continuation that did not adapt to it. As decoding narrows from $p=1$ to greedy, the two relations move apart. The NLL of the generated text falls from $3.5$ to $1.2$, while the increase at the first human token rises from $8.7$ to $9.6$. Random tokens give the contrast: despite an NLL of $13.6$, they raise the NLL of the first human token by only $6.4$, less than generated text under any decoding. SmolLM2 on FineWeb-Edu places random, sampled and greedy text in the same order (Appendix~\ref{app:generated-spans}). Narrower decoding therefore makes the generated trajectory easier to predict while making it harder to return to the fixed human continuation.

We next test whether this difference appears in training. Using top-p sampling with $p=0.9$ and greedy decoding on GPT-2 and top-p sampling on SmolLM2, we train on generated and random text under the routing conditions of Equation~\ref{eq:routing}. Relative to random tokens at the same positions, generated text causes less damage under target-only training and more under context-only training (Figure~\ref{fig:generated-spans}b). This holds for greedy and top-p generation on GPT-2 and for top-p generation on SmolLM2, where the two differences are about $0.05$ each.

Together, these results reveal a difference between how generated text is produced and learned from. Autoregressive generation is unusual as a source of training data because the same mechanism used to learn from text is also used to produce the text to be learned from. In observed text, a token is constrained by both what precedes it and what follows. Generated text is exempt from this constraint: \emph{it never has to fit its own continuation, because its continuation is generated to fit it}. Using autoregressive generation for training data therefore removes one of the constraints under which observed text is formed: compatibility with a continuation that does not adapt to the token.

\subsection{Routing the same position labels to the target or the context}
\label{sec:interfaces}

We next consider how to intervene on individual tokens during training. To isolate where an intervention should act, we provide oracle position labels, and take the corrupted positions of predictable replacement as known exactly. This leaves the question of what to do once those positions are known. The token at each position can be acted on through its own loss, where it is a target, or through the later predictions that read it as context. We compare the two with the same position labels. The comparison uses GPT-2 on OpenWebText and SmolLM2 on FineWeb-Edu (Appendix~\ref{app:operators}), and measures each intervention by the fraction of the damage under standard noise that it recovers.

\begin{figure}[t]
\vskip -0.05in
\centering
\begin{subfigure}[t]{0.535\textwidth}\centering\includegraphics[width=\linewidth]{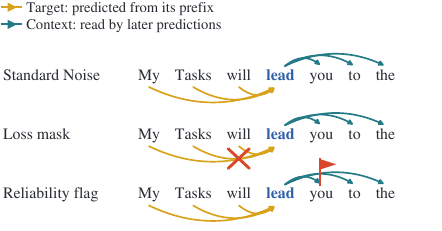}\caption{What each intervention changes}\end{subfigure}\hfill
\begin{subfigure}[t]{0.415\textwidth}\centering\includegraphics[width=\linewidth]{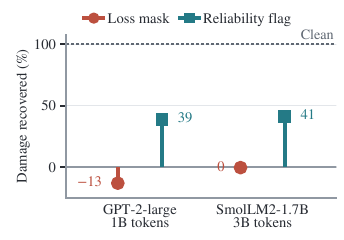}\caption{Damage recovered}\end{subfigure}
\vskip -0.05in
\caption{(a) The loss mask and the reliability flag use the same position labels in different roles, shown on the example of Figure~\ref{fig:32a}. Yellow arrows mark the prediction of the replaced token \emph{lead} from its prefix, and teal arrows the later predictions that read it. The mask removes the loss on \emph{lead} and leaves it readable; the flag keeps the loss and marks \emph{lead} for every later prediction that reads it. (b) At the oracle positions, the loss mask recovers none of the damage of predictable replacement and the flag about $40\%$. Damage recovered is the fraction of the standard noise damage removed, against each system's own clean and standard noise runs.}
\label{fig:oracle-interfaces}
\label{fig:remedies}
\vskip -0.1in
\end{figure}

\textbf{Loss masking.}
We begin with the intervention that existing token-level methods would take, removing the losses at the known positions~\citep{lin2024rho1,hans2024goldfish,wang2025slung}. Let $z$ be the observed stream and $m_i=1$ mark a corrupted position. The masked objective is
\begin{equation}
\mathcal{L}_{\mathrm{mask}}(z,m)=\sum_i(1-m_i)\,\ell_i(z_i\mid z_{<i}).
\label{eq:loss-mask}
\end{equation}
In the terms of Equation~\ref{eq:routing}, the mask removes the noisy token from the target stream and leaves it in the context stream. In the example of Figure~\ref{fig:32a}, the model is no longer trained to predict \emph{lead} after ``My Tasks will'', but it still reads \emph{lead} when it predicts ``you to the tasks'' (Figure~\ref{fig:oracle-interfaces}a). Unlike context-only routing, which supplies the clean target at a corrupted position, the mask supplies no target there, so it also removes about a tenth of all targets. The mask recovers none of the damage: the masked model ends slightly worse than the unmasked one on GPT-2 and level with it on SmolLM2, recovering $-13\%$ and $-0.3\%$ (Figure~\ref{fig:oracle-interfaces}b), and on GPT-2 it is better on only $17\%$ of the held-out windows. Knowing the right positions is therefore not enough when the intervention acts only on their target losses. The failure has two possible causes: the position labels may be of no use against this corruption, or the mask may deliver them to the wrong role. The latter is plausible because predictable replacement does more damage as context than as a target (Section~\ref{sec:comparison}), while masking its own loss leaves every later prediction to read it as before.

\textbf{Reliability flag.}
To distinguish the two explanations, we route the same labels to the context instead. Tags in the input already mark the quality or origin of training text~\citep{korbak2023pretraining,caswell2019tagged,allenzhu2024physics33}; we apply the same idea to single tokens. We keep every token and every loss and add the same label to the token's input representation,
\begin{equation}
h_i^{(0)}=e(z_i)+s\,r_{m_i},
\label{eq:reliability-flag}
\end{equation}
where $r_0,r_1$ are learned embeddings and $s$ fixes their strength. Because $z_i$ is predicted before it is read, its flag reaches only the predictions that use $z_i$ as context. The flag thus leaves the noisy token in both streams and adds to the context stream a mark of which tokens are unreliable: the model is still trained to predict \emph{lead}, and every later prediction reads \emph{lead} together with its mark (Figure~\ref{fig:oracle-interfaces}a). At evaluation every position is marked reliable. With $s=0.5$, the flag recovers $39\%$ of the damage on GPT-2 and $41\%$ on SmolLM2 (Figure~\ref{fig:oracle-interfaces}b), and on GPT-2 it lowers the loss on $97\%$ of the held-out windows. The same position labels therefore help once later predictions are allowed to read them.

\section{Conclusion}
\label{sec:conclusion}
Autoregressive pretraining gives every training token two roles: the model learns to predict it as a target and to use it as context. By separating the two roles, we find that they can favor opposite tokens and follow different relations to the surrounding text. The loss at a position therefore captures only part of what the model learns from the token there. Generated text makes the limitation concrete: generation produces text that is exceptionally easy to predict without ensuring that it is equally reliable to learn from as context. Token-level control likewise requires choosing which role to act through. More broadly, learning from data at the token level requires accounting separately for learning a token and learning from it.

\clearpage

\subsection*{AI use statement}

Language models and AI agents were used for coding assistance, grammar polishing, and phrasing refinement during the drafting of this paper. While artificial intelligence assisted in the presentation, all remaining errors, oversights, and bugs remain the exclusive product of natural stupidity.

\subsection*{Ethics statement}

This work investigates foundational training dynamics and data-intervention mechanisms in autoregressive language models using publicly available pretraining corpora. Our study does not involve human subjects, private user data, or hazardous capabilities.

\subsection*{Reproducibility statement}
Experiments use GPT-2~\citep{radford2019language}, SmolLM2~\citep{allal2025smollm2}, and Qwen3~\citep{yang2025qwen3} on OpenWebText~\citep{gokaslan2019openwebtext} and FineWeb-Edu~\citep{lozhkov2024fineweb-edu}. GPT-2 models are trained with nanoGPT~\citep{Karpathy2022}, SmolLM2 models with Nanotron~\citep{huggingface2023nanotron}, and Qwen3 models with Hugging Face Transformers~\citep{wolf2020transformers}. All experiments run on NVIDIA GH200 GPUs, with at most four per job. We report model sizes, training budgets, corruption rates, and multi-seed results in the main text, and provide complete training, corruption, and evaluation settings with the submission.

{\small\bibliography{lnt_references}\bibliographystyle{iclr2027_conference}}
\clearpage
\appendix
\section{Experimental Setup}
\label{app:setup}

Every trained model is evaluated with clean context and clean targets on held-out text that is disjoint, at the document level, from its training stream.

\subsection{GPT-2-large on OpenWebText}
\label{app:gpt2-setup}
This is the setup for the GPT-2 training comparison in Figure~\ref{fig1} and the held-out-window analysis in Figure~\ref{fig:relations}a.
The model has $773.5$M parameters, $36$ layers, $20$ heads, hidden size $1{,}280$, vocabulary size $50{,}257$, and no bias or dropout. Training uses nanoGPT, bf16 compute, sequence length $1{,}024$, and $491{,}520$ predicted tokens per update. AdamW has $(\beta_1,\beta_2)=(0.9,0.95)$, weight decay $0.1$, and gradient clipping $1.0$. The peak learning rate is $2.5\times10^{-4}$ with $1{,}000$ warm-up updates and cosine decay to $2.5\times10^{-5}$. The $9$B schedule has $18{,}311$ updates; all role arms are compared at update $18{,}250$, or $8{,}970{,}240{,}000$ predicted tokens. The $1$B runs of Section~\ref{sec:interfaces} use the same architecture and recipe and are compared at update $2{,}000$, or $983{,}040{,}000$ predicted tokens.

\subsection{SmolLM2 on FineWeb-Edu}
\label{app:smol-setup}
This is the setup for the SmolLM2 comparisons in Figure~\ref{fig:settings} and the four-stream experiment in Figure~\ref{fig:relations}b.
SmolLM2-1.7B has $24$ layers, hidden size $2{,}048$, feed-forward size $8{,}192$, $32$ attention heads and a tied vocabulary of $49{,}152$. It is initialized from scratch and trained with Nanotron in bf16, at sequence length $2{,}048$ and $256$ sequences per update ($524{,}288$ predicted tokens). AdamW uses $(0.9,0.95)$, $\epsilon=10^{-8}$, weight decay $0.1$ and gradient clipping $1.0$. The $3$B schedule has $5{,}722$ updates, a peak learning rate of $2.5\times10^{-4}$, $1{,}000$ warm-up updates, and cosine decay to $2.5\times10^{-5}$. The $1$B family is trained separately with the same recipe and ends at update $1{,}907$. The $10$B schedule reuses the architecture, initialization, data order and corruption definitions, extends the cosine schedule to $19{,}064$ updates, and consumes the full training stream without wrapping. The $135$M and $360$M models follow the $3$B recipe. The flag comparison of Section~\ref{sec:interfaces} uses $5{,}719$ updates ($2{,}998{,}403{,}072$ tokens) without wrapping and its own references (Appendix~\ref{app:operators}).

\subsection{Qwen3 continued pretraining}
\label{app:cpt-recipe}
This is the setup for the Qwen3 continued-pretraining comparisons in Figure~\ref{fig:settings}.
Qwen3-4B-Base is continued-pretrained on FineWeb-Edu for $1{,}000$ updates, with $512$ sequences per global update, $2{,}048$ predicted tokens per sequence, and $1.048576$B predicted tokens in total. The fixed stream contains $512{,}000$ non-overlapping chunks of $2{,}049$ stored tokens and is not wrapped. Paired role arms share a frozen chunk permutation and replacement realization. Training uses fp32 master parameters and AdamW states, bf16 forward compute, fp32 reductions, $(\beta_1,\beta_2)=(0.9,0.95)$, $\epsilon=10^{-8}$, weight decay $0.1$ on two-dimensional matrices and clipping $1.0$. A $5\%$ rewarm is followed by cosine decay to $10\%$ of the peak learning rate. The peak, $4\times10^{-5}$, is the largest of $10^{-5}$, $2\times10^{-5}$ and $4\times10^{-5}$ for which a $250$-step probe on clean text kept the held-out cross entropy within $0.01$ of the pretrained model's. Qwen3-1.7B-Base follows the same recipe. Predictable replacements are supplied by Qwen3-0.6B-Base from candidate ranks $1$--$41$, which gives them the same separation from the original tokens under the pretrained model's own loss as ranks $10$--$50$ give under a model trained from scratch.

\subsection{Corruption}
\label{app:corruption}
The two replacements are built to differ in prediction difficulty and in nothing else that we control. Random replacement draws uniformly from the vocabulary, so the replacement bears no relation to the prefix. Predictable replacement takes a frozen small language model's distribution over the next token given the clean prefix and draws the replacement from the candidates ranked $10$ to $50$. The top ranks are excluded because they often contain the original token itself or a close variant of it, which would leave the text uncorrupted; ranks $10$--$50$ give tokens that fit the prefix well but are still wrong, so their own loss lies between that of the original token and that of a random one (Figure~\ref{fig:32b}). Neither replacement sees the continuation, so any difference in fit to the continuation arises by chance.

Candidates for predictable replacement pass a complete-word filter. They must be space-initial and re-encode as a single token, so that the replacement occupies exactly the token it replaces and leaves the neighbouring tokens intact; alphabetic, at least four characters long and outside the stopword set, so that it is a content word rather than punctuation, a number or a function word; and inside a frequency-filtered English word list, so that it is a real word. The teacher is a pretrained GPT-2 (124M) for GPT-2 and SmolLM2-135M for SmolLM2, in each case given up to $512$ clean tokens of left context. On SmolLM2 the word list contains $13{,}756$ entries, and on average $27.6$ of the $41$ ranked candidates pass the filter, with a mean sampled rank of $30.5$. Positions at which no candidate passes are left unchanged, so at a requested rate of $10\%$ the realized rate is $9.766\%$ for predictable replacement and $9.990\%$ for random replacement.

\subsection{Constructing the four relation streams}
\label{app:relation-map}
To separate the two relations, we construct four replacement streams that cross low and high fit to the prefix with low and high fit to the continuation. For each candidate, a frozen $135$M teacher measures fit to the prefix by its log probability under the clean prefix. The same teacher then measures fit to the continuation by its mean loss on the next $16$ unchanged tokens after the candidate is inserted; lower continuation loss means better fit.

At each position, we restrict candidates to a common token-frequency quintile. We take four accepted candidates from the top of the teacher's ranking and four accepted candidates drawn uniformly, giving two levels of fit to the prefix. Within each group, the candidates with the lowest and highest continuation loss supply the two continuation conditions. This produces the four cells of the relation map. The two scores are nearly uncorrelated within a position (median Spearman correlation $-0.09$ in a $128$-position calibration), which allows the two relations to vary separately. Positions that cannot supply all four cells are skipped; fewer than $0.004\%$ of positions are skipped in any shard.

All four streams use the same $102{,}437{,}240$ replacement positions, a realized rate of $9.764\%$. Each stream is used once under target-only routing and once under context-only routing, giving eight training runs. Every run uses model seed $42$ and follows the $3$B-token SmolLM2-1.7B schedule for $2{,}000$ updates, or $1.048576$B training tokens.

\subsection{Own loss under a clean model}
\label{app:own-loss}
We first check that changing fit to the prefix changes the replacement's own prediction loss. A clean SmolLM2-1.7B checkpoint trained for $3$B tokens scores the original and replacement tokens under the same clean prefix. We evaluate all four relation-map streams at the same $10{,}000$ positions, spread evenly through the corruption stream and chosen so that their $2{,}048$-token prefixes do not overlap. At position $i$, we define the own-loss shift as
\[
\delta_i=-\log p_\theta(\tilde{x}_i\mid x_{<i})+\log p_\theta(x_i\mid x_{<i}).
\]
This is the NLL of the replacement minus that of the original token, so positive values mean that the replacement is harder to predict. Figure~\ref{fig:relations}b reports the mean shift for each stream.

\subsection{Where the original replacements fall}
\label{app:family-axes}
\begin{figure}[t]
\centering
\begin{subfigure}[t]{0.48\textwidth}\centering\includegraphics[width=\linewidth]{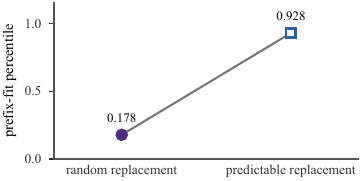}\caption{Fit to the prefix}\end{subfigure}\hfill
\begin{subfigure}[t]{0.48\textwidth}\centering\includegraphics[width=\linewidth]{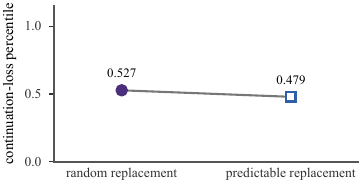}\caption{Fit to the continuation}\end{subfigure}
\caption{Where the original Random and Predictable replacement families fall on the two relations. Values are mean within-position percentiles over $128$ positions from each family.}
\label{fig:family-axes}
\end{figure}
The four streams above vary the two relations deliberately. We also locate the original Random and Predictable replacement families on the same axes. At $128$ sampled positions from each family, the frozen $135$M teacher scores fit to the clean prefix, and the clean SmolLM2-1.7B checkpoint trained for $3$B tokens scores the unchanged next $16$ tokens after each replacement. Because score scales differ across positions, we rank each realized replacement against alternative candidates at the same position and report its percentile. Predictable replacements lie near the top for fit to the prefix, with a mean percentile of $0.928$, whereas random replacements lie near the bottom, at $0.178$. Their continuation-loss percentiles are both near the middle, $0.479$ and $0.527$, and the bootstrap interval for their difference includes zero (Figure~\ref{fig:family-axes}). The original families are therefore strongly separated by fit to the prefix, but neither was selected for fit to the continuation.

\section{Generated Text}
\label{app:generated-spans}

This section gives the experimental details for Section~\ref{sec:generated}. The fixed-model scores correspond to Figure~\ref{fig:generated-spans}a, while the training settings and Standard Noise results correspond to Figure~\ref{fig:generated-spans}b.

\subsection{Data and decoding}
The GPT-2 experiments use a document-disjoint OpenWebText slice with $1.10$B GPT-2 tokens for training and the next $20$M tokens for validation. Starting at offset $17$, the training stream is partitioned into stretches of $32$ consecutive tokens; seed $1337$ selects $10\%$ of them without replacement, giving $3{,}437{,}497$ stretches and $109{,}999{,}904$ replaced positions. All corruption conditions use these positions and the same clean stream. A frozen pretrained GPT-2-medium receives up to $512$ clean tokens to the left of each stretch and generates exactly $32$ tokens, each conditioned on the tokens generated before it. End-of-text is suppressed during generation. Top-p sampling uses temperature $1$, $p=0.9$, and no top-$k$ truncation. Greedy decoding takes the top-ranked token at every step. The random condition uses one seed-$2024$ uniform-vocabulary realization at the same positions. The original tokens outside each stretch, including the continuation on its right, are unchanged.

The GPT-2 training runs in Figure~\ref{fig:generated-spans}b use randomly initialized GPT-2-medium models with context length $1{,}024$, a global batch of $262{,}144$ tokens, and $4{,}000$ updates ($1.048576$B tokens). AdamW uses $(\beta_1,\beta_2)=(0.9,0.95)$, weight decay $0.1$, clipping $1.0$, a $100$-update linear warmup to $3\times10^{-4}$, and cosine decay to $3\times10^{-5}$. Runs within each comparison share the model seed, data order, positions, corruption realization, and evaluation procedure.

The SmolLM2 experiments use the SmolLM2 tokenizer and a document-disjoint FineWeb-Edu slice containing $1{,}099{,}997{,}390$ training tokens and $19{,}999{,}369$ validation tokens. A frozen SmolLM2-135M generates sampled text with $p=0.9$ at the same kind of positions and, for scoring only, greedy text; the random condition uses the same positions. The training runs in Figure~\ref{fig:generated-spans}b use randomly initialized SmolLM2-135M models trained for $2{,}000$ updates of $524{,}288$ tokens ($1.048576$B tokens) with the recipe of Appendix~\ref{app:smol-setup}.

\subsection{Fixed-model scores: Figure~\ref{fig:generated-spans}a}
We select $20{,}000$ positions deterministically, excluding stretches that contain end-of-text, and score the replaced tokens and the first human token after them with a frozen model $p_\phi$, the generator itself. For a stretch starting at position $k$, the fit to the prefix is the mean NLL of the replaced tokens,
\[
-\frac{1}{32}\sum_{t=k}^{k+31}\log p_\phi(\tilde{x}_t\mid\tilde{x}_{<t}),
\]
and the fit to the continuation is the increase in NLL of the first human token,
\[
-\log p_\phi(x_{k+32}\mid\tilde{x}_{<k+32})+\log p_\phi(x_{k+32}\mid x_{<k+32}).
\]
Every sampled condition uses the same positions, prefixes, and sampling seed and generates $32$ tokens; only $p$ changes.

\subsection{Standard Noise: Figure~\ref{fig:generated-spans}b}
Under Standard Noise on GPT-2, the damage is $0.025$ and $0.028$ for random tokens in the greedy and sampled campaigns, $0.018$ for greedy text and $0.009$ for sampled text.

\section{Loss Mask and Reliability Flag}
\label{app:operators}
This section gives the settings for the single-token Predictable-replacement interventions in Section~\ref{sec:interfaces} and Figure~\ref{fig:oracle-interfaces}b.

Damage recovered is $(V_{\mathrm{std}}-V_{\mathrm{op}})/(V_{\mathrm{std}}-V_{\mathrm{clean}})$, where $V$ is held-out cross entropy and each system is compared with its own clean and Standard Noise runs; a negative value means that the intervention ends worse than Standard Noise.

\paragraph{GPT-2.}
The runs are compared at update $2{,}000$ of the $1$B recipe (Appendix~\ref{app:gpt2-setup}). The reliability flag adds a two-state embedding, initialized to zero, to the plain backbone of the clean and Standard Noise runs. The same flag placed at randomly chosen positions at the same rate recovers none of the damage ($-1\%$), so the gain of the oracle flag comes from the positions it marks rather than from the added embedding.

\paragraph{SmolLM2.}
Every run, including the clean and Standard Noise references, uses the backbone with the flag embedding, trained for $5{,}719$ updates of the $3$B recipe (Appendix~\ref{app:smol-setup}); evaluation sets every flag to the reliable state. Because that state still adds the learned embedding $s\,r_0$ to every token, and because creating the embedding changes the initialization of the backbone, these runs form their own model family, and interventions are compared only within it.

\end{document}